\documentclass[a4paper, 10pt, conference]{ieeeconf}      
\usepackage{cite} 
\usepackage{graphics} 
\usepackage{epsfig} 
\usepackage{mathptmx} 
\usepackage{times} 
\usepackage{amsmath} 
\usepackage{amssymb}  
\usepackage{longtable}
\usepackage{lscape}
\usepackage{multirow}
\usepackage{subfigure}
\usepackage{array}
\usepackage{hyperref}
\usepackage{url}
\usepackage{lipsum}
\usepackage{mathtools}
\usepackage{cuted}
\usepackage{FG2020}

\FGfinalcopy 

\IEEEoverridecommandlockouts                              
\def\FGPaperID{14} 

\title{\LARGE \bf
SAMM Long Videos: A Spontaneous Facial Micro- and Macro-Expressions Dataset
}

\author{\parbox{16cm}{\centering
    {\large Chuin Hong Yap, Connah Kendrick and Moi Hoon Yap}\\
    {\normalsize
    Department of Computing and Mathematics, Manchester Metropolitan University, Manchester, M1 5GD, UK\\
}
}
}

\begin{document}

\ifFGfinal
\thispagestyle{empty}
\pagestyle{empty}
\else
\author{Anonymous MEGC 2020 submission\\ Paper ID 14 \\}
\pagestyle{plain}
\fi
\maketitle

\begin{abstract}

With the growth of popularity of facial micro-expressions in recent years, the demand for long videos with micro- and macro-expressions remains high. Extended from SAMM, a micro-expressions dataset released in 2016, this paper presents SAMM Long Videos dataset for spontaneous micro- and macro-expressions recognition and spotting. SAMM Long Videos dataset consists of 147 long videos with 343 macro-expressions and 159 micro-expressions. The dataset is FACS-coded with detailed Action Units (AUs). We compare our dataset with Chinese Academy of Sciences Macro-Expressions and Micro-Expressions (CAS(ME)$^2$) dataset, which is the only available fully annotated dataset with micro- and macro-expressions. Furthermore, we preprocess the long videos using OpenFace, which includes face alignment and detection of facial AUs. We conduct facial expression spotting using this dataset and compare it with the baseline of MEGC III. Our spotting method outperformed the baseline result with F1-score of 0.3299.

\end{abstract}

\section{INTRODUCTION}
Facial expression research is multidisciplinary with various applications, such as emotional study, behavioural psychology, security ~\cite{Ekman_2009}, well-being ~\cite{Endres_Laidlaw_2009} and communication. In general, macro-expression refers to normal facial expression and micro-expression refers to a brief facial expression with duration of less than 500ms~\cite{Ekman_Friesen_1969}. Due to its involuntary nature, it is an important cue in non-verbal communication. 

In recent years, an international collaboration initiated by researchers ~\cite{yap2018facial, see2019megc} has conducted workshops and challenges on datasets and methods for facial micro-expressions recognition and spotting \cite{li2019spotting}. However, in real-world scenario, the occurrences of micro- and macro-expressions could co-exist or occur in isolation. Therefore, spotting micro- and macro-expressions is a challenging task. To date, there is limited Facial Action Coding System (FACS) coded long videos dataset. This paper presents SAMM Long Videos dataset, which consists of FACS coded micro- and macro-expressions. We release the dataset and its annotation for research uses.

The rest of the paper is organised as follows:
Section \ref{sec:background} describes the related work. Section~\ref{sec:samm} presents new analysis of the dataset, dataset preprocessing stage and performance metrics. Section~\ref{sec:result} discusses about the results of AU classification and spotting using this dataset. Section~\ref{sec:conclusion} concludes the paper. 


\section{Related Work}
\label{sec:background}

CAS(ME)$^2$ dataset~\cite{qu2017cas} is the only fully annotated dataset with both macro- and micro-expressions. There are 22 subjects and 87 long videos (in part A). The average duration is 148s. The facial movements are classified as macro- and micro-expressions. The video samples may contain multiple macro- or micro-expressions. The onset, apex and offset of these expressions were annotated and presented in a spreadsheet. Also, the eye blinks are labelled with the onset and offset frame. For MEGC II competition \cite{see2019megc} in 2019, the authors of SAMM \cite{davison2018samm} released 79 long videos for the micro-expressions grand challenge, but with only micro-expressions annotated and was only made available for MEGC II.

\begin{figure*}
	\centering
	\includegraphics[width=0.95\textwidth]{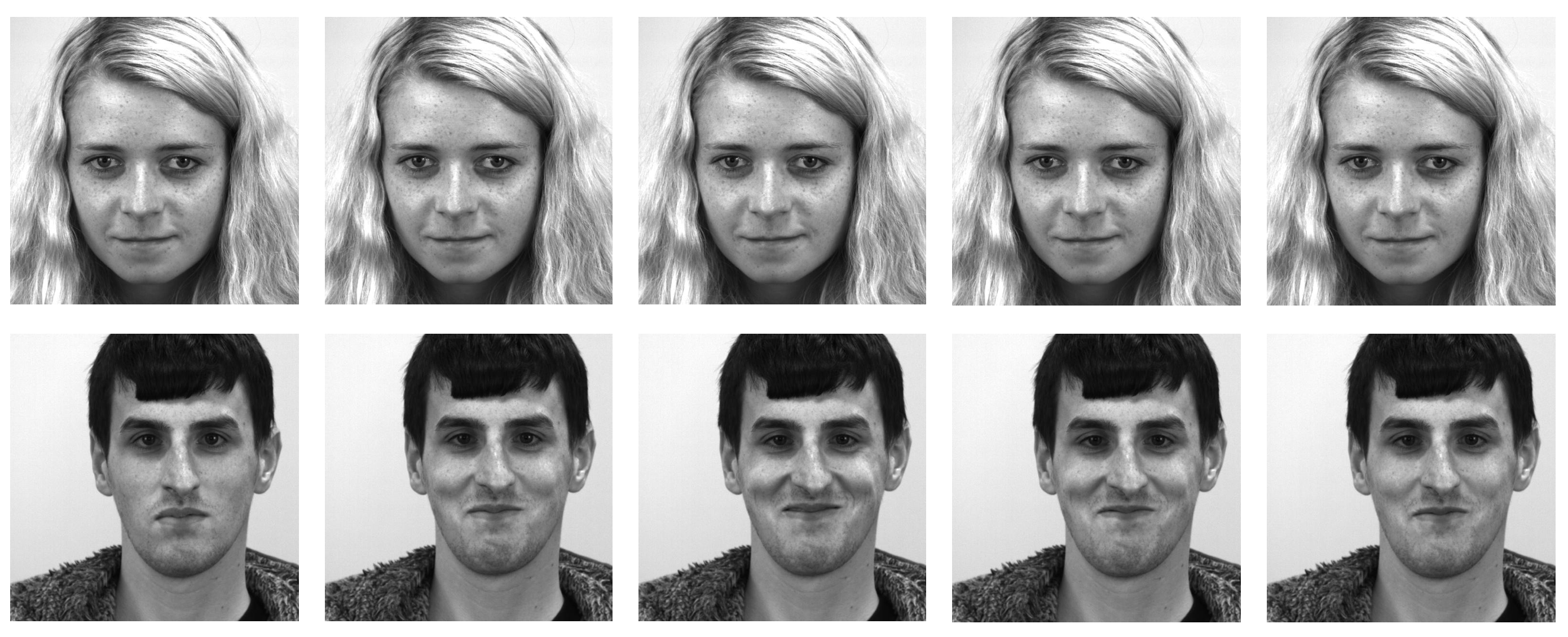}
	\caption{Two examples of facial expressions from SAMM Long Videos dataset: (Top) Micro-expression; and (Bottom) Macro-expression. Both expressions with AU12. On apex frame (middle), the macro-expression shows higher intensity and visibility when compared to the micro-expression.}
	\label{fig:sample}
\end{figure*}

In MEGC II, Li et al. \cite{li2019spotting} proposed temporal pattern extracted from local region for micro-expression spotting on two recently published datasets, i.e. SAMM~\cite{davison2018samm} (79 long videos) and CAS(ME)$^2$~\cite{qu2017cas}. Even though CAS(ME)$^2$ is labelled with macro- and micro-expressions, the focus of the challenge is on micro-expressions spotting. Li et al. \cite{li2019spotting} demonstrated their Local Temporal Pattern (LTP) method outperformed Local Binary Pattern (LBP) approaches, LBP-$\chi ^{2}$-distance, by Moilanen et al. ~\cite{Moilanen_Zhao_Pietikainen_2014}.

\section{Dataset Profile and Evaluation}
\label{sec:samm}
This section describes the specification of SAMM Long Videos, data preprocessing steps, facial AUs detection using OpenFace, facial movements detection method, and performance metrics for evaluation.

\subsection{Experiment} 
The original version of SAMM dataset~\cite{davison2018samm} is intended for facial micro-movements detection, which consists of a total of 32 subjects and has 7 videos each. The average length of videos is 35.3s. The original release of SAMM consisted of micro-movement clips with AUs annotated. Recently, the authors \cite{Davison2018} introduced objective and emotion classes for the dataset. In 2018, MEGC I used the objective classes for the recognition challenge. In 2019, MEGC II's recognition challenge used the emotional classes from the database as ground truth. In addition to recognition challenge, the spotting challenge was introduced. However, due to the size of SAMM Long Videos dataset, the organisers only used a subset of 79 long videos, each contains one or more micro-movements, with a total of 159 micro-expressions. The index of onset, apex and offset frames of micro-movements were provided as the ground truth. Although the long videos were released for the grand challenge, the macro-expressions labels were not provided.

The micro-movements interval is defined from onset to offset frame. In this dataset, all the micro- and macro-movements are labelled. Thus, the spotted frames can indicate not only facial expressions but also other facial movements, such as eye blinks. The details of the experimental settings, eliciting process and coding process are described in Davison et al. \cite{Davison2018}.

\subsection{Comparison of Facial Expressions and Dataset}
Figure \ref{fig:sample} shows two examples of facial expressions in SAMM Long Videos. The top row illustrates a micro-expression of brief AU12 with low intensity and the bottom row shows a macro-expression of AU12 with high intensity.

When compared to CAS(ME)$^2$, SAMM has more long videos. The resolution and frame rate of SAMM is higher than CAS(ME)$^2$. When compared to the number of facial expressions, SAMM has 159 micro-expressions (which is similar to the first release by Davison et al. \cite{davison2018samm}) and 343 macro-expressions (newly FACS-coded macro-expressions for this release). Table \ref{tab:comparison} compares the differences between SAMM and CAS(ME)$^2$.

\begin{table}[h]
	\caption{A Comparison between SAMM Long Videos and CAS(ME)$^2$}
	\centering
	\begin{tabular}{|c||c|c|}
		\hline
		Dataset & SAMM Long Videos & CAS(ME)$^2$  \\
		\hline
		Number of Long Videos &147 &87\\
		\hline
		Number of Videos with micro &79 &32\\
		\hline
		 Resolution &2040$\times$1088 &640$\times$480\\
		\hline
		Frame rate &200 &30\\
		\hline
		Number of Macro-expressions &343 &300\\
		\hline
		Number of Micro-expressions&159 &57\\
		\hline
	\end{tabular}
	
	\label{tab:comparison}
\end{table}


\begin{figure}
	\centering
	\includegraphics[width=.5\textwidth]{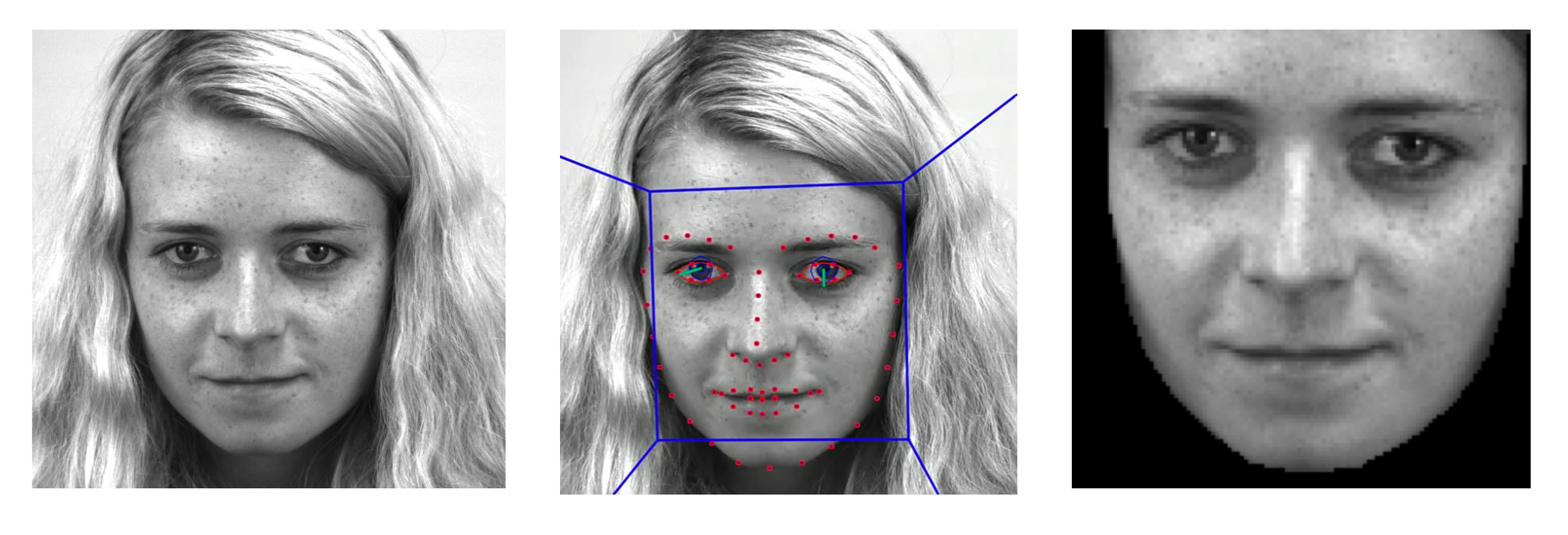}
	\caption{An illustration of preprocessing steps using OpenFace: (Left) Original SAMM image; (Middle) Facial Landmark Detection; and (Right) Cropped face ROI.}
	\label{fig:preprocess}
\end{figure}

\subsection{Dataset Preprocessing}
\label{sec:method}

We preprocess SAMM Long Videos using OpenFace which includes facial landmark detection, facial expression recognition, head pose and eye gaze estimation. For our case, we focus on face alignment and detection of AUs only.

\subsubsection{Face alignment}
OpenFace \cite{baltruvsaitis2016openface, baltrusaitis2018openface} is a general-purpose toolbox for face recognition, which consists of face alignment algorithm using affine transformation. The facial landmarks are detected by Dlib's face landmark detector \cite{king2009dlib}. Figure \ref{fig:preprocess} illustrates the original SAMM image, the facial landmarks and the aligned face image.
OpenFace uses a similarity transform on current detected landmarks to a representation of landmarks from a neutral expression. This process maps the face texture to a common reference frame and removes changes due to scaling and in plane rotation. The output image has a dimension of $112\times112$ pixel with interpupilary distance of 45 pixels.

OpenFace utilises Convolutional Experts Constrained Local Model (CE-CLM) which uses deep networks to detect and track facial landmark features.
The deep network was simplified from the original version which contained 180,000 parameters to around 90,000. This reduces the model size and speeds up the model by 1.5 times with minimal accuracy loss.
Moreover, CE-CLM uses 11 initialisation hypotheses at different orientations which lead to 4 times performance improvement. It also utilises sparse response maps which further improved the model speed by 1.5 times.

\begin{figure*}
	\centering
	\includegraphics[width=0.95\textwidth]{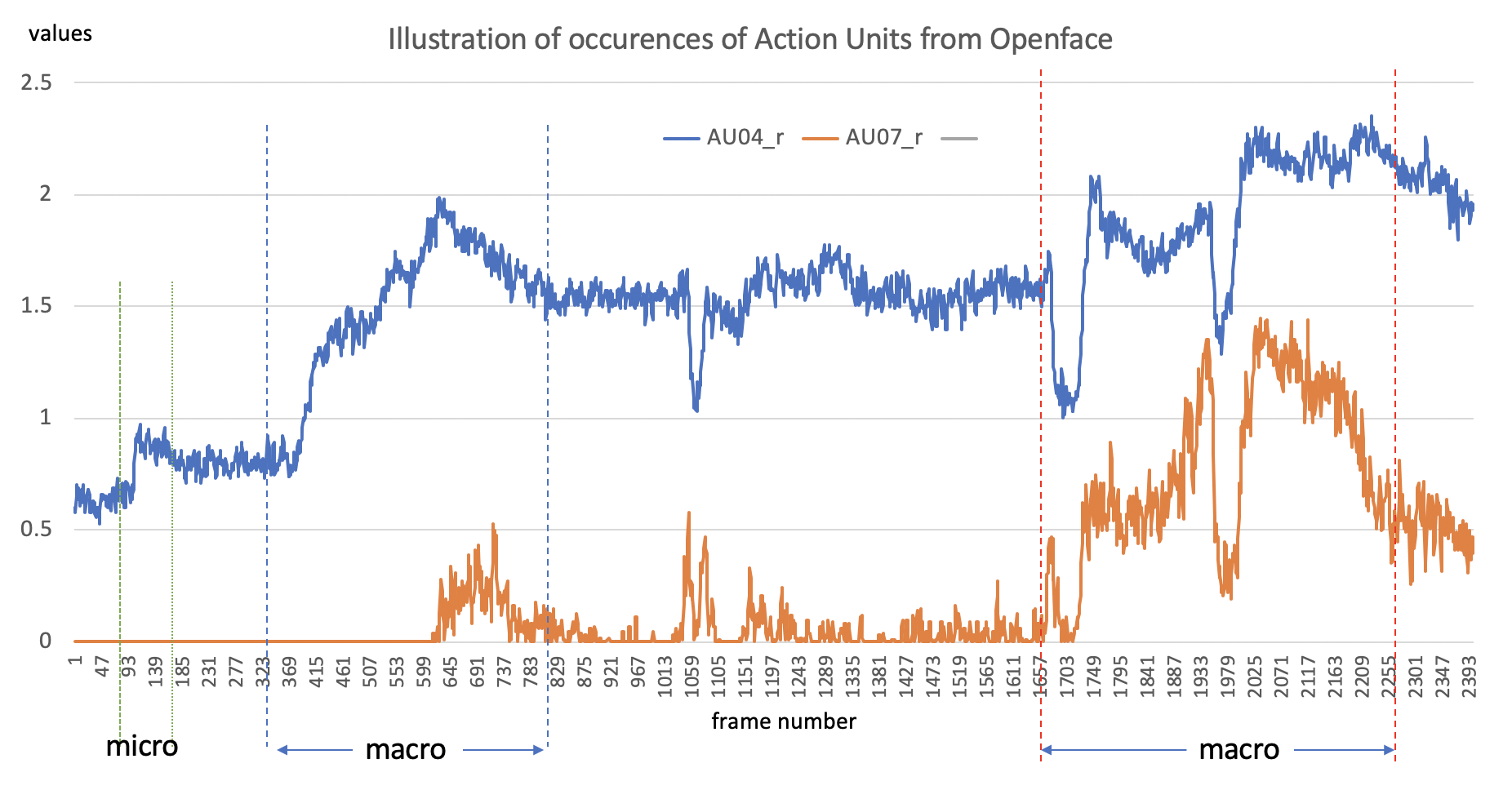}
	\caption{Two AUs on long video clips, the blue line is AU4, and the orange line is AU7. There are one micro-expression and two macro-expressions found on this video clip.}
	\label{fig:AU_samples}
\end{figure*}

\subsubsection{Detection of AUs}
OpenFace \cite{baltruvsaitis2016openface, baltrusaitis2018openface} is capable of detecting the presence and intensity of AUs. Figure \ref{fig:AU_samples} illustrates the sample output of OpenFace on two AUs plotted on a graph, where there are one micro-expression and two macro-expressions in the video clip. 
OpenFace conducts facial expression recognition through detecting the intensity and presence of AUs. The intensity of AU is represented on a 5-point scale while the presence of AU is encoded 0 as absent and 1 as present. The AUs that are capable to be recognised by OpenFace are 1, 2, 4, 5, 6, 7, 9, 10, 12, 14, 15, 17, 20, 23, 25, 26, 28, and 45. The full name of the facial part represented by each AUs are shown in Table \ref{tab:listofAU}. Linear kernel Support Vector Machines was used to detect AU occurrence while Support Vector Regression was use to detect the intensity of AU\cite{baltruvsaitis2015cross}. It also uses the concatenation of Histogram of Oriented Gradient (HOG) of the dimension reduced face image and facial shape features obtained from CE-CLM for each facial features.
Note that AU45 involves detecting the presence of eye blink. It was removed from the output signal as it does not carry any significant information in facial expressions classification and spotting.


\begin{table}[h]
	\caption{List of AUs in OpenFace}
	\centering
	\begin{tabular}{|c|c|}
		\hline
		\textbf{AU} & \textbf{Full name}\\
		\hline
		AU1 & Inner Brow Raiser\\
		\hline
		AU2 &  Outer Brow Raiser \\
		\hline
		AU4 &  Brow Lowerer \\
		\hline
		AU5 &  Upper Lid Raiser \\
		\hline
		AU6 &  Cheek Raiser \\
		\hline
		AU7 &  Lid Tightener \\
		\hline
		AU9 &  Nose Wrinkler \\
		\hline
		AU10 &  Upper Lip Raiser \\
		\hline
		AU12 &  Lip Corner Puller \\
		\hline
		AU14 &  Dimpler \\
		\hline
		AU15 &  Lip Corner Depressor \\
		\hline
		AU17 &  Chin Raiser \\
		\hline
		AU20 &  Lip Stretcher \\
		\hline
		AU23 &  Lip Tightener \\
		\hline
		AU25 &  Lips Part \\
		\hline
		AU26 &  Jaw Drop \\
		\hline
		AU28 &  Lip Suck \\
		\hline
		AU45 &  Blink\\
		
		\hline
	\end{tabular}
	
	\label{tab:listofAU}
\end{table}

\subsection{Detection of Facial Movements}
\subsubsection{Signal smoothing}
\begin{figure*}
	\centering
	\includegraphics[width=1\textwidth]{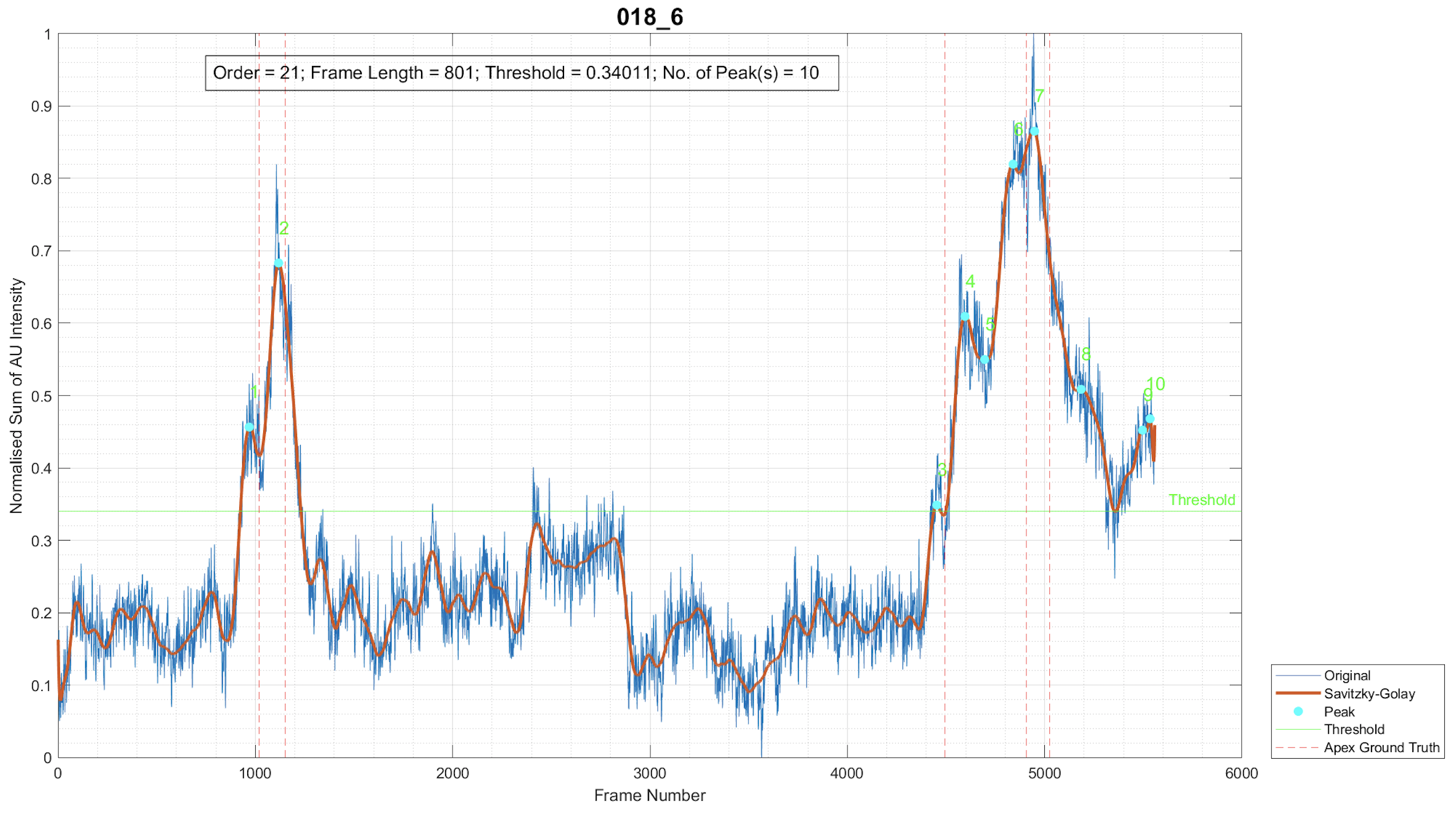}
	\caption{Plot of Subject 018\_6 for the normalised sum of AU intensity (blue line) and the filtered signal (red line) with peaks annotated. A threshold value of 75\textsuperscript{th} percentile of normalised AU intensity was used to filter out the peaks originated from noise.}
	\label{fig:AU_patterns}
\end{figure*}

For spotting task, we combine the intensity of all the detected AUs, and normalised it to a scale of 0 to 1. Subsequently, Savitzky-Golay filter \cite{savitzky1964smoothing} with 21\textsuperscript{st} order was used for noise reduction. Figure \ref{fig:AU_patterns} illustrates a subject with the normalised sum of the intensity of AU and the smoothed signal plotted with respect to frame number.

\subsubsection{Onset and offset frame acquisition}
By using a custom analysis algorithm modified from \cite{hewitt2018novel}, the onset and offset frame of the processed data were obtained and compared with the ground truth. This algorithm uses Daubechies wavelet \cite{daubechies1990wavelet} with scaling function of 2 and level 3 signal decomposition for signal smoothing. For peak detection, peak was defined as a point where it is higher than the 7 points that come immediate before and after it. A threshold value of 75\textsuperscript{th} percentile of normalised AU intensity were also implemented. This filters out low local peaks originated from noise. The spotted interval is obtained by subtracting the onset from the offset. Since the frame rate of SAMM Long Videos is 200fps, a threshold of 100 frames (0.5s) was selected to classify the spotted intervals into macro-expressions ($>$0.5s) or micro-expressions ($\leq$0.5s).

\subsection{Performance Metrics}
Since micro- and macro-expressions occur over a series of frames, the measurement (accuracy) based on the individual frame is not widely accepted in this domain. The preferable performance metrics used are based on the overlap of the frames (based on the concept of Intersect over Union (IoU) in computer vision). Following the spotting challenge in \cite{li2019spotting}, a true positive (TP) is defined as 

\begin{equation}
\frac{Predicted\cap GT}{Predicted\cup GT} \geq k
\end{equation}
where $k$ is set to 0.5, $GT$ represents the ground truth expression interval (onset-offset), and $Predicted$ represents the detected expression interval.
Otherwise, the spotted interval is regarded as \textit{false positive (FP)}. \textit{False negative (FN)} occurs when the ground truth expression interval exists in the absence of predicted interval. 

Then, the comparison of the spotting accuracy will be measured in $Precision$, $Recall$, and $F1-Score$ as in equation (2), equation (3) and equation (4), respectively.

\begin{equation}
Precision = \frac{TP}{TP+FP}
\end{equation}

\begin{equation}
Recall = \frac{TP}{TP+FN}
\end{equation}

\begin{equation}
F1-score = \frac{2TP}{2TP+FP+FN}
\end{equation}

For AUs detection, we report the $Accuracy$ of our results which is the number of AUs correctly detected by OpenFace divided by the total number of AUs in ground truth.

\begin{table}[h]
	\caption{Accuracy of AU Classification of OpenFace}
	\centering
	\begin{tabular}{|c|c|c|c|}
		\hline
		& Micro & Macro & Combined \\
		\hline
		Matched AU& 80 & 245 & 278\\
		\hline
		Ground Truth & 172 & 501 & 590\\
		\hline
		Accuracy & 0.4651 & \textbf{0.4890} & 0.4712\\
		\hline
	\end{tabular}
	
	\label{tab:classification}
\end{table}

\begin{table}[h]
	\caption{Comparison between AU classification of videos containing only micro- or macro-expression}
	\centering
	\begin{tabular}{|c|c|c|}
		\hline
		& Micro only videos & Macro only videos\\
		\hline
		Matched AU& 9 & 104\\
		\hline
		Ground Truth & 31 & 231 \\
		\hline
		Accuracy & 0.2903 & \textbf{0.4502}\\
		\hline
	\end{tabular}
	
	\label{tab:macromicro}
\end{table}
\begin{table*}[htbp]
	\caption{Results of macro- and micro-spotting in SAMM Long Videos compared to the baseline of MEGC III.  }
	\centering
	\begin{tabular}{|c|c|c|c|c|c|c|c|}
		\hline
		& \multicolumn{3}{c|}{\textbf{Our Results}}& \multicolumn{3}{c|}{\textbf{Baseline of MEGC III}\cite{he2019spotting}} \\
		\hline
		Expression &macro-expression &micro-expression &overall &macro-expression &micro-expression &overall \\
		\hline
		Total number &343             &159             &502          &343 &159 &502  \\
		\hline
		TP           &\textbf{172}    &6               &\textbf{178} & 22 & \textbf{29} & 51\\
		\hline
		FP           &328             &71              &399          & 334 & 1407 & 1741\\
		\hline
		FN           &171             &153             &324          & 321 & 130 & 451 \\
		\hline
		Precision   &\textbf{0.3440}  &\textbf{0.0779} &\textbf{0.3085} & 0.0618 & 0.0202 & 0.0285\\
		\hline
		Recall      &\textbf{0.5015}  &0.0377          &\textbf{0.3546} & 0.0641 & \textbf{0.1824} & 0.1016\\
		\hline
		F1-score    &\textbf{0.4081}  &\textbf{0.0508} &\textbf{0.3299} & 0.0629 & 0.0364 & 0.0445\\
		\hline
	\end{tabular}
	\label{tab:result}
\end{table*}

\section{Result and Discussion}
\label{sec:result}


For AU classification, AU presence in each video were compared with ground truth. This measures the ability of OpenFace to detect and distinguish AUs. The $Accuracy$ for AU classification is shown in Table \ref{tab:classification}.

OpenFace does not classify AU presence by macro- or micro-expression and we observe that there are 84 overlapped AUs (between macro- and micro-expressions). 
As a result, the detected AU in these two classes contains \textit{false positives} and \textit{false negatives} from the overlapped AUs. To compare the performance of OpenFace on micro- and macro-movements detection, we analyse the micro-only videos and macro-only videos. The comparison between the performance metric of videos containing only macro- or micro-expression is shown in Table \ref{tab:macromicro}. 
All videos used in this comparison contains either with macro- or micro-expressions which gives a fairer evaluation in the detection accuracy of these two classes. There are 18 micro-only and 68 macro-only videos in this dataset. As we observe, micro-expression detection has a lower performance compared to macro-expression detection. This indicates that OpenFace has lower accuracy in detecting AU of micro-expression.

For facial movements detection, the results and comparison with baseline results \cite{he2019spotting} are shown in Table \ref{tab:result}. 
A sorting algorithm were used to combine consecutive detection of TPs into a single measurement. As a result, it was found that there are misclassification of micro-expressions as a single macro-expression. A check on the TPs of micro- and macro-expression was also performed and found that there are 43 micro- misclassified as macro-expressions and 9 macro- misclassified as micro-expressions.
All of them were labelled as FPs and FNs. Moreover, by following the criteria set by MEGC III spotting challenge, the overlapping between the prediction and ground truth interval of the TPs found was calculated using equation (1). It was found that 93 TPs did not fulfil the benchmark and they were labelled as FPs and FNs as well.
Overall, the sorting algorithm works well as we had verified that the total number of ground truth facial expressions matched the sum of TPs and FNs.

In Table \ref{tab:result}, it is shown that our spotting method outperformed the baseline. In both results, macro-expression spotting yields the highest F1-score. This confirms that macro-expression spotting is easier compared to micro-expression spotting. It can also be explained as macro-expression has longer interval ($>0.5$s) and hence easier to be spotted by the spotting algorithm. Moreover, OpenFace was trained on facial expressions datasets, which have no labelled micro-expressions in the training set, which explain the low F1-score in spotting micro-expressions.




\par
\section{Conclusion}
\label{sec:conclusion}
This paper presents SAMM Long Videos dataset on facial micro- and macro-expressions. We evaluated the performance of OpenFace facial behaviour tools in AUs detection. As OpenFace was not designed to detect facial micro-expressions, it achieved a reasonable results with $Accuracy$ of 0.4712 and $F1-Score$ of 0.3553. It performs poorly in micro-only videos with $Accuracy$ of 0.2903 in contrast to 0.4502 in macro-only videos.
For spotting, our overall results outperformed the baseline result of MEGC III in all three performance metrics of $Precision$, $Recall$, and $F1-Score$.

SAMM Long Videos will benefit researchers in many disciplines, such as affective computing, human behavioural, computer vision and machine learning community. 

Future work will extend the dataset by improving class balance by including long videos that contain more micro-expressions. More effective method of elicitation of micro-expression should be investigated.

SAMM Long Videos dataset is publicly available at http://www2.docm.mmu.ac.uk/STAFF/M.Yap/dataset.php.

\section{Acknowledgement}
This project was supported by Manchester Metropolitan University Vice-Chancellor’s PhD Studentship. The authors gratefully acknowledge the contribution of FACS Coders and collaborators in Davison et al. \cite{davison2018samm}.


\bibliographystyle{IEEEtran}
\bibliography{IEEEabrv,reference}

\end{document}